\pdfoutput=1 

\documentclass[runningheads]{llncs}

\usepackage{eccv}

\usepackage{eccvabbrv}

\usepackage{graphicx}
\usepackage{booktabs}

\usepackage[accsupp]{axessibility}

\usepackage{hyperref}

\usepackage{orcidlink}
\usepackage{bm}

\begin{document}

\title{Rethinking Sparse Lexical Representations for Image Retrieval in the Age of Rising Multi-Modal Large Language Models}

\titlerunning{Rethinking Sparse Lexical Representations for Image Retrieval}

\author{Kengo Nakata \and
Daisuke Miyashita \and
Youyang Ng \and
\\
Yasuto Hoshi \and
Jun Deguchi}

\authorrunning{K.~Nakata et al.}


\institute{
 Kioxia Corporation, Yokohama, Japan \\
\email{\{kengo1.nakata, daisuke1.miyashita, youyang.ng, \\ yasuto1.hoshi, jun.deguchi\}@kioxia.com}}
\maketitle

\begin{abstract}
In this paper, we rethink sparse lexical representations for image retrieval.
By utilizing multi-modal large language models (M-LLMs) that support visual prompting, we can extract image features and convert them into textual data,
enabling us to utilize efficient sparse retrieval algorithms employed in natural language processing for image retrieval tasks.
To assist the M-LLM in extracting image features, we apply data augmentation techniques for key expansion and analyze the impact with a metric for relevance between images and textual data.
We empirically show the superior precision and recall performance of our image retrieval method compared to conventional vision-language model-based methods on the MS-COCO, PASCAL VOC,
and NUS-WIDE datasets in a keyword-based image retrieval scenario, where keywords serve as search queries.
We also demonstrate that the retrieval performance can be improved by iteratively incorporating keywords into search queries.

\end{abstract}

\section{Introduction}
\label{sec:intro}

As deep learning technologies have evolved,
deep neural networks (DNNs) have achieved exceptional performance in image recognition and object detection tasks~\cite{AlexNet,ResNet,FasterRCNN,MaskRCNN}, 
and approaches leveraging these networks have been extensively explored for image retrieval tasks~\cite{DeepImageRetrieval,DynamicMatchCNN,UnsupervisedPart-basedWeighting,InstanceLevelRetrievalTransformers}.
With the recent emergence and widespread adoption of vision-language models~\cite{CLIP,ALIGN,BLIP,CrossModalImageSegment},
text-to-image retrieval has become one of the mainstream research areas in image retrieval.
These models are pre-trained on vast amounts of paired image-text data collected from the internet,
and they learn to map the images and their corresponding texts into similar dense vector representations in a shared latent space~\cite{CLIP,ALIGN,BLIP}.
By utilizing such pre-trained models, images that are semantically similar or related to a query text can be retrieved based on the distance calculations between their dense vectors.

Images can contain a wide variety of information and features,
and the criteria to determine whether images are similar or dissimilar are inherently subjective and not uniquely defined.
For instance, when retrieving images, the features that users focus on within images may vary depending on individual preferences and situational factors.
However, a query does not always represent or reflect a user's desires or intentions, and it could be incomplete or lacking in required information to specify them.
Despite these limitations, the vision-language model attempts to provide results that are relevant to the user's request,
by implicitly compensating for the lack of information in the incomplete query based on the knowledge acquired through training.
Unfortunately, this compensated information may not always align with the user's desires or intentions.

In text retrieval, a keyword-based approach is commonly employed in practical applications to retrieve documents containing specified keywords.
Users can combine multiple keywords as search queries to specify their focus areas or topics. 
After viewing the retrieval results, users can iteratively refine their search queries by modifying and/or adding keywords as feedback.
Even if the retrieval model cannot initially provide the desired results, users can adaptively obtain results that align with their preferences or intentions through this iterative process.
In the light of these flexible capabilities, we aim to explore better methods for applying keyword-based retrieval to image retrieval tasks.

While we can directly apply the conventional vision-language models to the keyword-based retrieval, 
we cannot overlook the substantial advancements in large language models (LLMs) over the past few years,
which have demonstrated a remarkable ability to comprehend context within dialogue interfaces~\cite{GPT3,GPT4,BlenderBot3,LLaMA,LLaMA2,LaMDA}.
Additionally, multi-modal LLMs (M-LLMs) have already been proposed to comprehend visual information within images through visual prompting,
which involve processing images along with textual data as queries~\cite{BLIP2,InstructBLIP,TagGPT,LENS,LLaVA,LLaVA15}.
By utilizing the advanced capabilities of M-LLMs, we can extract features from images and linguistically represent them in textual data like tags and captions.
Then, we can leverage the advantages of natural language processing (NLP) techniques for image retrieval tasks.
We encode the generated textual data into sparse lexical vectors and utilize efficient retrieval algorithms to enable effective image retrieval based on their sparse lexical representations.

In this paper, we focus on the text-to-image retrieval task and rethink the task in this age of rising such powerful M-LLMs.
As a text-to-image retrieval task, we consider a keyword-based image retrieval scenario where a search query consists of a few words representing the contents or objects depicted in images. 
Through quantitative analysis on the benchmark datasets,
we demonstrate that our retrieval system outperforms conventional vision-language model-based retrieval methods in terms of precision and recall.
Specifically, we introduce a cropping technique to assist the M-LLM in effectively extracting image features, and analyze the effectiveness by evaluating a metric for relevance between images and texts.
As our findings, we empirically show that the conventional vision-language model-based methods outperform our approach, if a less informative caption is used as a search query.
This seems to depend on whether there is a function to compensate for the lack of information in the less informative query.
However, the retrieval performance of our system improves significantly when we incorporate keywords into the search query, making the query explicitly informative.

The main contributions of this paper are as follows:
\begin{itemize}
\item We introduce a text-to-image retrieval system that utilizes M-LLMs and retrieval algorithms based on sparse lexical representations, and evaluate its effectiveness on various benchmark datasets.
\item To enhance M-LLM performance in extracting image features, we employ data augmentation techniques for key expansion and quantitatively evaluate the improvement in retrieval performance.
\end{itemize}

\section{Related Work}
\subsection{Evolution of Image Retrieval Research}

Image retrieval has been studied extensively in recent decades.
This includes tasks such as finding images similar to a given input image,
searching for images with specific content features like colors, shapes, and textures~\cite{CBIR_2D,CBIR_SVM},
and retrieving images based on their semantic meaning or content categories~\cite{SemanticRetrieval}. 
Researchers have also explored more specialized tasks like conditioned image retrieval,
where both images and semantic conditions are used as query inputs to find relevant images~\cite{EffectiveConditionedImageRetrieval,ConditionedImageRetrievalCLIP}.
In practice, image retrieval often involves searching for images based on descriptive texts,
such as metadata and hashtags utilized on social media platforms. This requires matching user queries with relevant images by analyzing their associated metadata or the images themselves.
However, despite its practical significance, text-to-image retrieval remains an understudied domain within computer vision research.
This lack of research may be attributed to the fact that the accuracy of the image retrieval heavily depends on the quality and thoroughness of human annotations.

Recent advances in NLP have empowered text-to-image retrieval through the development of contrastive language-image pre-training techniques~\cite{CLIP,ALIGN,BLIP}. 
By mapping images and texts into a shared latent space and calculating their semantic similarities,
these techniques enable image retrieval based on textual descriptions.
Moreover, the rise of M-LLMs that can generate textual descriptions for images without the need for human annotations~\cite{GPT4V_1,GPT4V_2,LLM_annotation}
emphasizes the significance of exploring interactions between images and descriptive texts in the form of interpretable lexical data.
In this paper, we rethink text-to-image retrieval by leveraging sparse lexical representations.
For the context of text-to-image retrieval tasks, while previous research has primarily focused on the evaluations in the caption-to-image retrieval~\cite{CLIP,ALIGN,ImageBERT,UNITER,BLIP,FLAVA},
we shift our attention to keyword-based image retrieval, which is more commonly used in practical applications but remains understudied.
We explore its potential applications and provide insights for future research directions.

\subsection{Large Language Models for Visual Promptings}
In the past few years, LLMs have made remarkable progress, and their popularity has grown significantly
due to their impressive ability to comprehend context~\cite{GPT3,GPT4,BlenderBot3,LLaMA,LLaMA2,LaMDA}.
Recently, there has been a growing development of M-LLMs, which support visual inputs as well as text-based promptings,
providing comprehensive and flexible applications~\cite{TagGPT,LENS,LLaVA,LLaVA15}.
GPT-4V is capable of accepting both text and image prompts, understanding visually depicted scenarios in images,
and addressing complex visual question answering tasks~\cite{GPT4,GPT4V_1,GPT4V_2}.
TagGPT offers a tagging system that extracts tags from multi-modal content such as images and videos
without requiring additional knowledge or human annotations, by leveraging the M-LLMs~\cite{TagGPT}.
BLIP-2 and InstructBLIP introduce Querying Transformer that bridges the modality gap between images and texts, while keeping pre-trained vision encoders and backbone LLMs frozen~\cite{BLIP2,InstructBLIP}.
LLaVA proposes a visual instruction tuning technique for large multi-modal models that integrate vision encoders with LLMs
for general-purpose visual and language understanding by utilizing instruction-following data generated by GPT-4~\cite{LLaVA,LLaVA15}.
The rise of these M-LLMs presents an opportunity for us to revisit the understudied keyword-based image retrieval without the need for human annotations. 
In this paper, we leverage these M-LLMs to transform visual information in images into expanded lexical representations, enabling us to harness traditional efficient sparse retrieval methods for image retrieval tasks.
Our approach effectively combines classic techniques with cutting-edge innovations.

\subsection{Sparse Retrieval}
Recently, there has been growing interest in using dense retrieval methods
that leverage dense vector representations generated
by DNNs for image retrieval tasks~\cite{DeepImageRetrieval,DynamicMatchCNN,UnsupervisedPart-basedWeighting,InstanceLevelRetrievalTransformers}.
However, sparse vector representations, typically in the form of lexical retrievers,
have also been explored due to their enhanced interpretability and analytical capabilities~\cite{LexLIP,STAIR,ImageSuggestion,DisentangledRepresentation}.
To address the perceived trade-off between accuracy and interpretability,
LexLIP~\cite{LexLIP} introduces a lexicon-weighting paradigm
to significantly reduce retrieval latency while maintaining high performance with bag-of-words models.
Similarly, STAIR~\cite{STAIR} maps images and texts to a sparse token space
to construct sparse text and image representations for improved retrieval accuracy. 
These studies demonstrate the potential of sparse retrievers to outperform dense retrievers.
Our approach leverages a multi-modal language model to extract image features into textual data.
We then utilize vectorization and retrieval algorithms in NLP tasks, such as BM25~\cite{BM25}, TF-IDF~\cite{TF_IDF}, and word2vec~\cite{word2vec}, for image retrieval tasks. 
Among these techniques, BM25 is considered an efficient sparse retrieval algorithm and is frequently used for benchmark evaluations in information retrieval tasks~\cite{DisentangledRepresentation,LexLIP,BRIGHT}.
BM25 demonstrates better out-of-distribution generalization capabilities compared to dense retrievers~\cite{BEIR},
and outperforms them in retrieving named entities or words that were not seen during training~\cite{ChallengeDenseRetrievers}.
Based on such potential capabilities, we employ the efficient and standard retrieval algorithm, BM25, for sparse lexical vectors directly converted from textual data.
Our approach does not rely on dense latent representations extracted by DNNs, and eliminates the need for specialized vector space adaptation,
enabling the application of key expansion techniques for enhanced performance.
By transforming the image retrieval task into a sparse lexical retrieval task, we can rethink image retrieval from an NLP perspective.

\section{Approach}
\begin{figure}[t!]
   \begin{center}
   \includegraphics[width=0.9\linewidth]{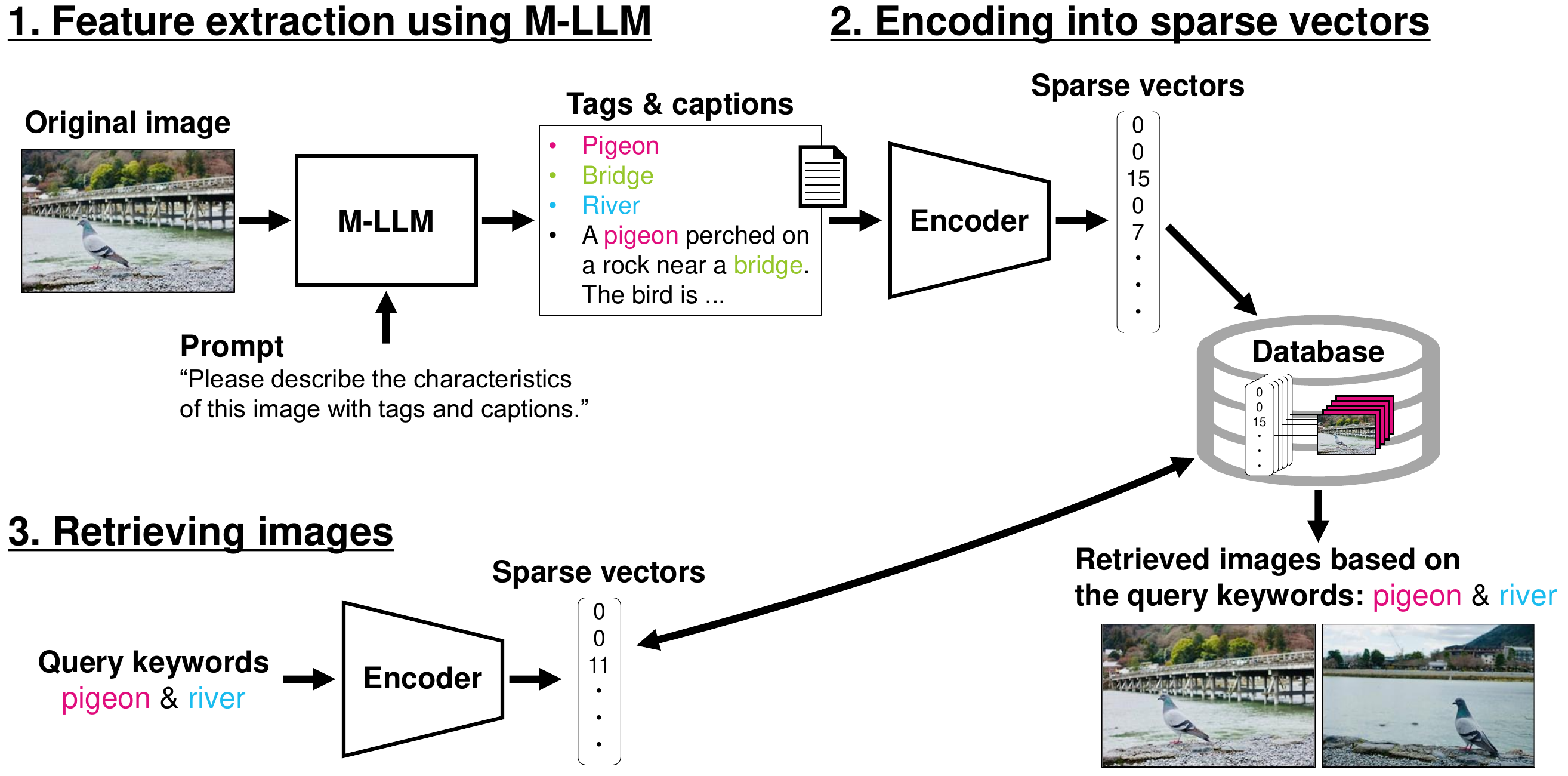} 
   \end{center}
   \caption{Overview of our image retrieval system.
    Our system utilizes an M-LLM to describe an image in textual data such as tags and captions.
    The textual data is encoded into sparse vectors. 
    When retrieving specific images, query keywords are also encoded into sparse vectors, enabling the retrieval of relevant images.}
   \label{fig:overview}
\end{figure}

Fig.~\ref{fig:overview} provides an overview of our image retrieval system.
Our image retrieval system consists of three processes:
(1) feature extraction using an M-LLM,
(2) encoding into sparse vectors, and
(3) retrieving images.
Our system accepts text-based queries such as keywords, and returns a set of relevant images from a database.
Each process is described below.

\subsection{Feature Extraction Using M-LLM}

First, we generate textual data for images to be stored in a database.
By utilizing M-LLMs with visual prompting capabilities~\cite{GPT4,LENS,TagGPT,LLaVA},
it is possible to extract features from images and represent them in textual data.
Among the M-LLMs, the pre-trained LLaVA model demonstrates high performance across various benchmark datasets for general-purpose visual and language understanding~\cite{LLaVA,LLaVA15},
therefore we utilize this publicly available model in our work.
We provide the M-LLM with an image and a prompt such as ``\texttt{Please generate multiple captions to describe the features of this image.}''
or ``\texttt{Please describe the characteristics of this image with tags and captions.}''
Then, we can obtain generated lexical tags and captions that represent the image features.

For our system, we can also apply image captioning models~\cite{ReflectiveDecoding,mPLUG,CoCa}.
However, we choose to utilize pre-trained M-LLMs provided in the open-source library, as they offer generally powerful performance without the need for model tuning~\cite{LLaVA,LLaVA15}.
The remarkable ability of M-LLMs to interactively comprehend prompts and context can be utilized to iteratively extract information related to image content by providing them with step-by-step queries.
For instance, after the M-LLM generates captions for an image based on the initial prompt, we provide it with another prompt:
``\texttt{If there are any additional features of this image that are not expressed in the generated captions, please generate additional captions to explain them.}''
Considering such potential applications, we present the system using M-LLMs instead of image captioning models in this paper.

\begin{figure}[t]
   \begin{center}
   \includegraphics[width=0.9\linewidth]{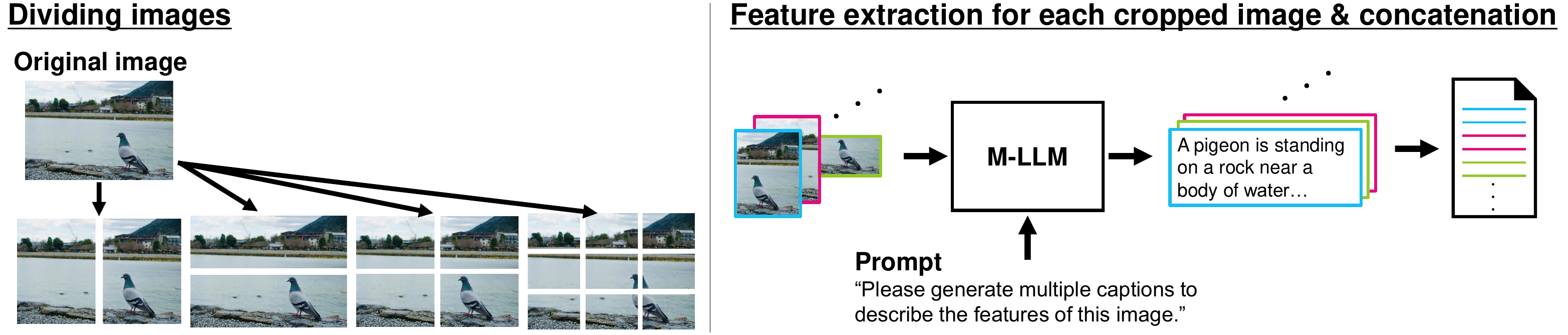} 
   \end{center}
   \caption{Data augmentation techniques for key expansion. An original image is segmented into multiple regions as cropped images (left),
   and each cropped image is processed by an M-LLM to generate captions that extract the features of each region (right). 
   By concatenating the generated captions, including those derived from the original image, we can extract a comprehensive set of features from the whole image.}
   \label{fig:cropping_images}
\end{figure}

\subsubsection{Data Augmentation Techniques for Key Expansion.}

In order to sufficiently extract features and information from various viewpoints within images,
we employ a cropping technique, which is commonly used in image recognition tasks as a means of data augmentation.
As shown in Fig.~\ref{fig:cropping_images}, we divide an original image into multiple segments as cropped images, such as two vertical segments, two horizontal segments, four segments, or nine segments.
For each cropped image, the M-LLM generates a corresponding caption that describes the feature of the image.
By concatenating all the generated captions, including those derived from the original image, we can extract a comprehensive set of features from the whole image.
The cropping technique assists the M-LLM in effectively extracting features from images, while expanding the key sets in the database (i.e., key expansion), leading to improved retrieval performance.

For cropping images, we can also utilize object detection models like YOLO or spatial transformer networks~\cite{YOLO,YOLOv3,SpatialTransformer}.
However, when using the object detection model, areas where the model fails to detect objects or where no objects exist
(e.g., sky scenery, glass fields, or sea areas) will not be cropped.
Consequently, the M-LLM cannot extract information from these areas within the images.
To avoid the impact of inductive biases in object detection models, in this paper, we do not use such models, but instead employ fixed pattern cropping,
regardless of the location of objects within the images, as shown in Fig.~\ref{fig:cropping_images}.

\subsubsection{Analysis with CLIPScore.}
\label{sec:analysis_clipscore}

To evaluate the impact of the cropping technique, we use a metric called CLIPScore~\cite{CLIPScore}.
CLIPScore is used to evaluate the relevance between an image and a textual description by comparing the embeddings extracted through the models pre-trained by CLIP.
If the pre-trained model embeds an image data ($I$) and a textual data ($T$) into their respective embeddings $\bm{E}_I$ and $\bm{E}_T$,
we can calculate CLIPScore based on the cosine similarity between their embeddings as follows,
\begin{align}
{\rm CLIPScore}(I, T) &= w \times {\rm max}({\rm cos}(\bm{E}_I, \bm{E}_T), 0), \label{eq:clip_score} \\
{\rm cos}(\bm{E}_I,\bm{E}_T) &= \frac{\bm{E}_I \cdot \bm{E}_T}{||\bm{E}_I||||\bm{E}_T||}, \label{eq:cos_sim}
\end{align}
where $w$ is a scaling parameter used to adjust the range of score distribution,
and we set $w = 2.5$ as reported in the original paper~\cite{CLIPScore}.
This metric is typically used for evaluating the quality of image captioning models, by measuring the relevance between the captions generated by the models and the corresponding images~\cite{CLIPScore,CLIPScore_1,CLIPScore_2}.
In contrast, we utilize this metric to evaluate the effectiveness of the cropping technique in eliciting diverse textual descriptions.
For example, if the CLIPScore value between a cropped image and a textual description is larger than the value between the original image and the textual description, 
it indicates that the cropping technique has produced a more suitable image for eliciting the textual description.

In our evaluation, we use the Microsoft COCO (MS-COCO) dataset~\cite{MSCOCO}, which comprises images depicting a variety of scenarios involving multiple objects from 80 different categories.
Specifically, we employ the 2017 validation set of MS-COCO, consisting of 5,000 images.
We adopt the list of 80 categories as diverse textual descriptions (e.g., ``bicycle'' and ``refrigerator''),
and calculate CLIPScore based on Eqs.~\eqref{eq:clip_score} and~\eqref{eq:cos_sim} between each image and textual description.
We utilize a model pre-trained by CLIP (\texttt{ViT-L/14@336px}) for vision and text encoders, as in the experiments described in Sec.~\ref{sec:experimental_setup}.
We average CLIPScore over all the images and all the textual descriptions as follows,
\begin{align}
{\rm AveragedCLIPScore_{all}} &= \frac{1}{N_{I}}\sum_{i=1}^{N_I}{\rm AveragedCLIPScore_{each}}(I_i), \label{eq:averaged_clipscore} \\
{\rm AveragedCLIPScore_{each}}(I_i) &= \frac{1}{(N_{C}+1)N_{T}}\sum_{j=0}^{N_C}\sum_{k=1}^{N_T}{\rm CLIPScore}(I_{i,j}, T_{k}), \label{eq:averaged_clipscore_each}
\end{align}
where $N_I$ is the total number of images in the dataset, while $N_C$ represents the number of cropped images for each image
and $N_T$ denotes the number of textual descriptions.
Additionally, $T_k$ represents the $k$-th textual description, and $I_{i,j}$ refers to the $j$-th cropped image for the $i$-th image.
Specifically, when $j=0$, it corresponds to the original image before cropping.

\begin{figure}[t]
   \begin{center}
   \includegraphics[width=0.9\linewidth]{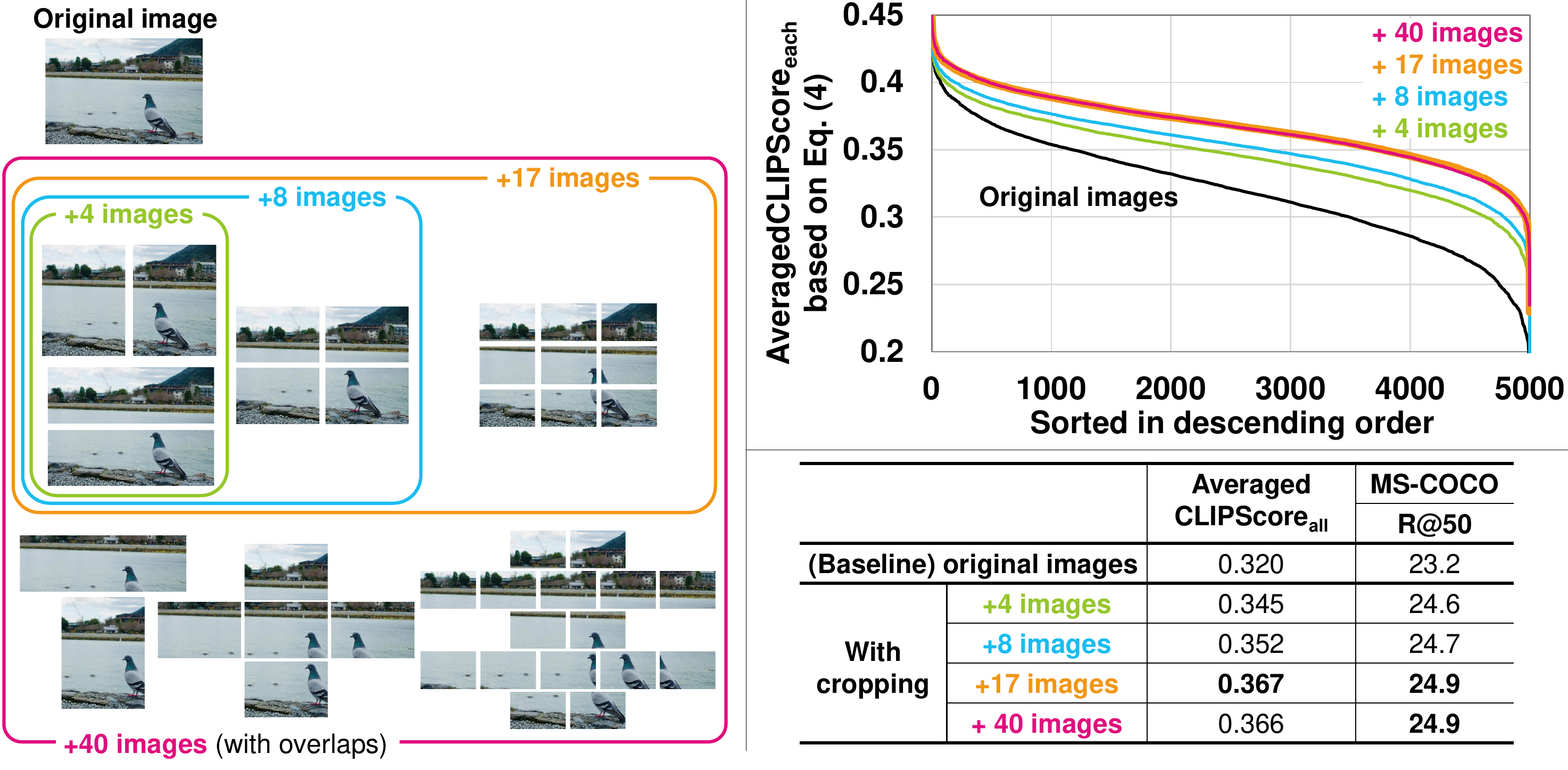} 
   \end{center}
   \caption{The variations in averaged CLIPScore based on Eq.~\eqref{eq:averaged_clipscore_each} for each of the 5,000 validation images from the MS-COCO dataset.
   As shown in the left figure, the original images are cropped by fixed patterns including overlaps.
   In the upper right graph, the values are sorted by averaged CLIPScore for each image in descending order.
   The lower right table summarizes averaged CLIPScore based on Eq.~\eqref{eq:averaged_clipscore} for all the images in the dataset, along with the top-50 recall performance (R@50).
   }
   \label{fig:clipscore}
\end{figure}

As shown in the left figure of Fig.~\ref{fig:clipscore}, the original images are cropped by fixed patterns, which include overlapped edges, in order to evaluate the impact of information loss along the boundaries of the cropped patterns.
The upper right figure in Fig.~\ref{fig:clipscore} shows the variations in averaged CLIPScore based on Eq.~\eqref{eq:averaged_clipscore_each}
for each of the 5,000 validation images from the MS-COCO dataset.
As the number of cropped images increases to 17, averaged CLIPScore based on Eq.~\eqref{eq:averaged_clipscore} for all the images in the dataset increases, as summarized in the lower right table.
We also evaluate the retrieval performance, which is measured by top-50 recall (R@50), in retrieving the images
(the recall is calculated as in Eq.~\eqref{eq:recall_k} by evaluating keyword-based image retrieval scenario described in Sec.~\ref{sec:experimental_setup}).
As the number of cropped images increases to 17, we can observe the improvement on R@50. 
On the other hand, when the number of cropped images increases from 17 to 40 by cropping with overlaps, averaged CLIPScore based on Eq.~\eqref{eq:averaged_clipscore} does not increase and the recall performance is not improved.
At this point, we consider that the impact of cropping has reached a point of saturation, and the information loss from the original images has been effectively mitigated.
Hence, we select 17 fixed patterns for cropping in the experiments of Sec.~\ref{sec:experiments}.

By employing the cropping technique, images can be divided into multiple segments, which reduces the number of features that the M-LLM needs to focus on and describe for a single image.
The above results empirically indicate that the cropping technique enhances the ability of the M-LLM to extract features from images more precisely, leading to improved retrieval performance.

\subsection{Encoding into Sparse Vectors and Text-to-Image Retrieval}

We encode the textual data generated by the M-LLM into lexical representations with sparse vectors,
where we insert non-zero values only at the positions that correspond to the terms present in the corpus.
We employ the BM25 algorithm~\cite{BM25} for effective image retrieval based on the sparse lexical representations.
BM25 is a widely used algorithm in NLP applications that efficiently retrieves documents
by scoring them based on their term frequencies,
enabling the search for relevant documents to a given query.
Specifically, this algorithm assigns higher weights to rare terms within the corpus and lower weights to common terms.
These vectors are then stored in the form of an inverted index,
allowing for quick lookups of documents containing specific terms and significantly reducing the search space,
thereby accelerating the retrieval process.

When searching for images, we set descriptive keywords as search queries to focus on specific features or aspects within images.
We convert these search queries into sparse lexical representations, and we can retrieve relevant textual data and corresponding images from the database based on their sparse lexical representations by using the BM25 algorithm.
The actual settings, such as parameters, are described in the subsequent section. 

\section{Experiments}
\label{sec:experiments}

\subsection{Experimental setup}
\label{sec:experimental_setup}

\subsubsection{Model settings.}
A LLaVA model is an end-to-end trained large multi-modal model that connects a vision encoder and an M-LLM for general-purpose visual and language understanding~\cite{LLaVA,LLaVA15}.
Compared to other M-LLMs with visual prompting capabilities, the LLaVA's large multi-modal model has demonstrated superior performance on a variety of benchmark datasets, outperforming models like BLIP-2 and InstructBLIP~\cite{LLaVA,LLaVA15}.
In our experiments, we utilize one of the pre-trained multi-modal models (\texttt{llava-1.5-13b-hf})\footnote{
We utilize the pre-trained models available at the following URLs: \\
\url{https://huggingface.co/llava-hf/llava-1.5-13b-hf}}
publicly available from the Hugging Face's Transformers library~\cite{TransformersLibrary}.
Based on our analysis in Sec.~\ref{sec:analysis_clipscore},
we divide original images to obtain 17 cropped images as shown in the left figure of Fig.~\ref{fig:clipscore}.
We provide the pre-trained LLaVA model with the prompt ``\texttt{Please generate multiple captions to describe the features of this image.}''
for each of the original images and the cropped images, in order to generate captions that represent the features of each image. 
After the caption generation, we concatenate all the generated captions.

We use zero-shot vision-language models 
pre-trained by CLIP (\texttt{ViT-L/}\\ \texttt{14@336px})\footnote{
\url{https://github.com/openai/CLIP/blob/main/clip/clip.py}} and ALIGN (\texttt{align-base})\footnote{\url{https://huggingface.co/kakaobrain/align-base}} as our baselines,
because these models have demonstrated robust and reliable performance across a wide range of benchmark datasets~\cite{CLIP,ALIGN}.
Note that the pre-trained LLaVA model utilizes the vision encoder model included in the same pre-trained CLIP model as its vision encoder.
Given this, we can expect that our chosen M-LLM has similar visual performance capabilities as one of our baseline models.
Neither our method nor the baseline methods involves fine-tuning the pre-trained models. 

\subsubsection{Task settings and datasets.}
As a text-to-image retrieval task, we consider a keyword-based image retrieval scenario using three benchmark datasets:
MS-COCO~\cite{MSCOCO}, PASCAL VOC~\cite{PascalVOC}, and NUS-WIDE~\cite{NUS-WIDE}.
Each dataset consists of images featuring multiple objects and scenes, and each image is annotated with descriptive labels.
In our experiments, we utilize the ground-truth labels assigned to each image in the respective datasets as keywords for our search queries to retrieve the corresponding images.
For instance, the 2017 validation set of MS-COCO contains a total of 5,000 images, of which 4,952 images include one or more objects belonging to 80 categories with 80 different label types.
Then, each of these 80 distinct labels like ``\texttt{bus}'' serves as a keyword for our search query.
Similarly, we utilize the 2007 test set of PASCAL VOC, which comprises 4,952 images with one or more objects per image belonging to 20 classes.
Moreover, we explore the NUS-WIDE dataset, featuring 260,648 web images with one or more textual tags per image.
Each image is labeled with multiple concepts from a set of 81 labels.
As a subset, we focus on the 195,834 image-text pairs corresponding to the 21 most common concepts,
and we use a total of 2,100 image-text pairs from this subset for the test set, as previously validated in~\cite{NUS-WIDE-tc21,NUS-WIDE-tc21_journal}.

We also consider a multi-keyword-based image retrieval scenario, where multiple keywords are combined and used to refine the search criteria like an AND search.
For our experiments, we join the ground-truth labels of each image in the aforementioned datasets into a search query.
For example, if we use an image file named ``\texttt{val2017/000000454661.jpg}'' that contains labeled objects such as ``\texttt{car}'', ``\texttt{bus}'', and ``\texttt{traffic light}'' in the MS-COCO dataset,
we join the labels together as ``\texttt{car}, \texttt{bus}, \texttt{traffic light}'' to form the search query for the image.

Furthermore, we evaluate the performance in a caption-to-image retrieval setting, which is a basic evaluation setting for text-to-image retrieval tasks.
We utilize the MS-COCO and Flickr30k~\cite{Flickr30k} datasets, and employ the ground-truth caption sentence for each image as a search query to retrieve the corresponding image.
Specifically, we use a total of 5,000 images in the 2017 validation set of MS-COCO and a total of 1,000 images in the test set of Flickr30k.

Finally, in order to explore the potential practical applications of our retrieval system,
we consider a scenario for text-to-image retrieval with user feedback.
In this scenario, after an initial retrieval based on a search query,
a user iteratively incorporates keywords into the search query as user feedback to gradually clarify the vision for the desired image like a multi-turn refined search.
As an example, we combine the caption-to-image retrieval setting with the keyword-based image retrieval setting.
We utilize both ground-truth captions and labels for 4,952 images in the 2017 validation set of MS-COCO.
After an initial retrieval based on a ground-truth caption for an image, we iteratively incorporate the ground-truth labels for the image into the search query as keyword-based user feedback.

\subsubsection{Retriever settings.}
In our retrieval system, we leverage the BM25 algorithm and employ Pyserini~\cite{Pyserini}, which is a Python interface to Lucene's BM25 implementation.
We set the parameters to their default values of $k_1 = 0.9$ for term frequency scaling and $b = 0.4$ for document length normalization.
In the baseline, the vision-language models retrieve images by calculating the distance based on cosine similarity between the embeddings for query texts and images.

In the keyword-based image retrieval evaluations,
we directly utilize ground-truth labels as textual inputs for the text encoder models.
We do not employ prompt templates like ``\texttt{A photo of a \{label\}.}'' commonly used in CLIP~\cite{CLIP}.
Our preliminary experiments showed that the use of such templates could potentially decrease recall performance,
particularly when the labeled object was not the main focus of the image.
Therefore, we do not use such templates and instead directly use the labels.

\subsubsection{Evaluation metrics.}
To evaluate the retrieval performance, we use precision and recall metrics.
We sweep the number of top-retrieved images ($k$) by powers of two from 1 to the total number of images in each dataset.
We calculate the precision ($P@k$) and recall ($R@k$) metrics based on the number of true positives ($TP@k$)
among the top $k$ retrieved images for each query ($q$) as follows,
\begin{align}
P@k &= \frac{\sum_{q=1}^{N_Q}TP_{q}@k}{N_{Q}k}, \label{eq:precision_k} \\
R@k &= \frac{\sum_{q=1}^{N_Q}TP_{q}@k}{\sum_{q=1}^{N_Q}P_q}, \label{eq:recall_k}
\end{align}
where $N_{Q}$ is the total number of queries and $P_q$ is the total number of ground-truth images for each query.

\begin{figure}[b!]
   \begin{center}
   \includegraphics[width=1.0\linewidth]{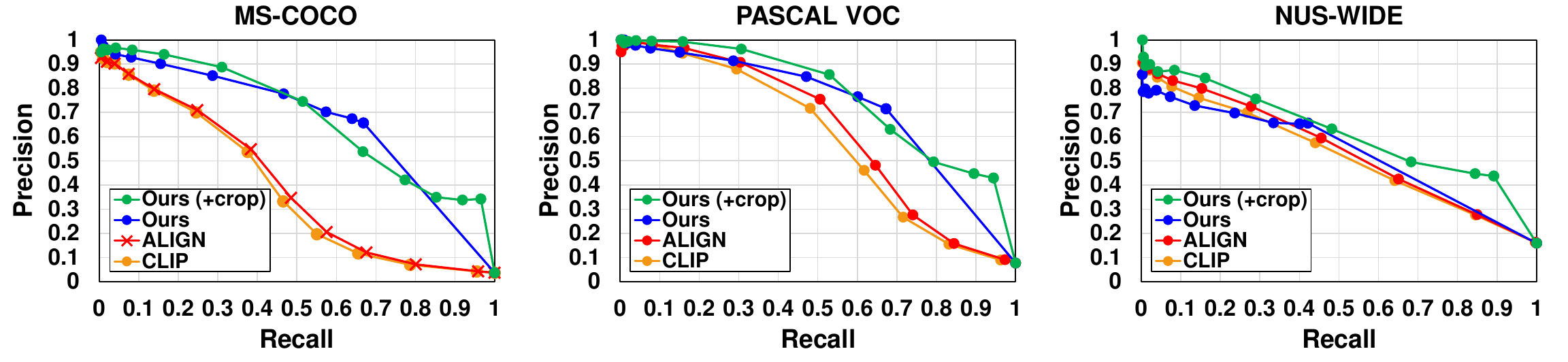} 
   \end{center}
   \caption{Comparison of precision and recall curves between our retrieval method and the conventional retrieval methods for the keyword-based image retrieval setting on the MS-COCO, PASCAL VOC, and NUS-WIDE datasets.}
   \label{fig:pr_curves_single_label_to_image_retrieval}
\end{figure}

\subsection{Experimental results}
\subsubsection{Keyword-based image retrieval.}
As shown in Fig.~\ref{fig:pr_curves_single_label_to_image_retrieval},
our retrieval method exhibits higher precision and recall compared to the conventional vision-language model-based methods
on the MS-COCO and PASCAL VOC datasets for the keyword-based image retrieval setting.
By using the cropping technique on the NUS-WIDE dataset,
we can observe an improvement in both precision and recall performance, outperforming the conventional methods.

\begin{figure}[t!]
   \begin{center}
   \includegraphics[width=1.0\linewidth]{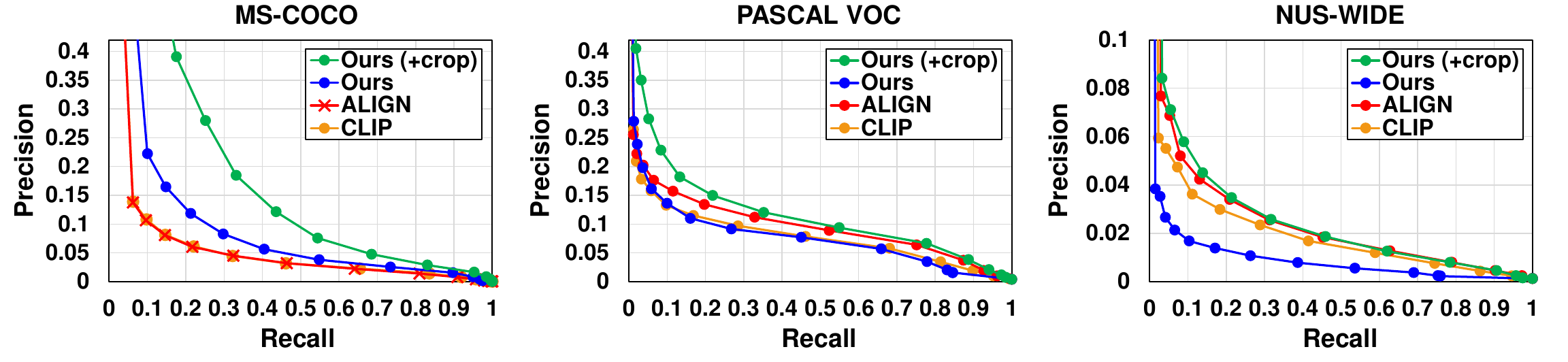} 
   \end{center}
   \caption{Comparison of precision and recall curves between our retrieval method and the conventional retrieval methods for the multi-keyword-based image retrieval setting on the MS-COCO, PASCAL VOC, and NUS-WIDE datasets.}
   \label{fig:pr_curves_multi_label_to_image_retrieval}
\end{figure}

Fig.~\ref{fig:pr_curves_multi_label_to_image_retrieval} shows the precision and recall curves in the multi-keyword-based image retrieval setting.
As shown in Fig.~\ref{fig:pr_curves_multi_label_to_image_retrieval}, both precision and recall are improved by using the cropping technique.
Specifically, our method outperforms the conventional methods by using the cropping technique on the PASCAL VOC and NUS-WIDE datasets.

To quantitatively compare the precision and recall performance, 
we summarize the values of PR-AUC (Area Under the Precision-Recall Curve) on the MS-COCO, PASCAL VOC, and NUS-WIDE datasets in Table~\ref{table:pr_auc_comparison}.
Our retrieval method outperforms the conventional methods on the MS-COCO dataset, demonstrating higher PR-AUC values.
In addition, the utilization of the cropping technique on the PASCAL VOC and NUS-WIDE datasets improves the PR-AUC values, surpassing the performance of conventional methods.

\begin{table}[t!]
\begin{center}
\caption{Comparison of PR-AUC values for the keyword-based image retrieval settings on the MS-COCO, PASCAL VOC, and NUS-WIDE datasets using our retrieval method and the conventional retrieval methods.
Multi indicates the PR-AUC values in the multi-keyword-based image retrieval setting.}
\scalebox{1.0}{
\begin{tabular}{p{5em}p{12em}p{12em}p{12em}p{12em}p{12em}p{12em}}
\hline 
&\multicolumn{2}{c}{\textbf{MS-COCO}}&\multicolumn{2}{c}{\textbf{PASCAL VOC}}&\multicolumn{2}{c}{\textbf{NUS-WIDE}}\\
&\multicolumn{2}{c}{2017 validation set}&\multicolumn{2}{c}{2007 test set}&\multicolumn{2}{c}{21 classes test set}\\ \cline{2-7}
\multicolumn{1}{c}{\textbf{Method}}&\multicolumn{1}{c}{}&\multicolumn{1}{c}{Multi}&\multicolumn{1}{c}{}&\multicolumn{1}{c}{Multi}&\multicolumn{1}{c}{}&\multicolumn{1}{c}{Multi}\\ \hline 
\multicolumn{1}{c}{\textbf{CLIP}~\cite{CLIP}}&\multicolumn{1}{c}{0.382}&\multicolumn{1}{c}{0.070}&\multicolumn{1}{c}{0.587}&\multicolumn{1}{c}{0.083}&\multicolumn{1}{c}{0.523}&\multicolumn{1}{c}{0.029}\\
\multicolumn{1}{c}{\textbf{ALIGN}~\cite{ALIGN}}&\multicolumn{1}{c}{0.398}&\multicolumn{1}{c}{0.069}&\multicolumn{1}{c}{0.622}&\multicolumn{1}{c}{0.100}&\multicolumn{1}{c}{0.543}&\multicolumn{1}{c}{0.036}\\ \hline 
\multicolumn{1}{c}{\textbf{Ours}}&\multicolumn{1}{c}{0.666}&\multicolumn{1}{c}{0.112}&\multicolumn{1}{c}{0.722}&\multicolumn{1}{c}{0.080}&\multicolumn{1}{c}{0.535}&\multicolumn{1}{c}{0.016}\\
\multicolumn{1}{c}{\textbf{+crop}}&\multicolumn{1}{c}{\textbf{0.682}}&\multicolumn{1}{c}{\textbf{0.210}}&\multicolumn{1}{c}{\textbf{0.765}}&\multicolumn{1}{c}{\textbf{0.123}}&\multicolumn{1}{c}{\textbf{0.619}}&\multicolumn{1}{c}{\textbf{0.039}}\\ \hline 
\end{tabular}
}
\label{table:pr_auc_comparison}
\end{center}
\end{table}

\subsubsection{Caption-to-image retrieval.}

In Table~\ref{table:recall_comparison_caption}, we summarize the recall performance for the caption-to-image retrieval setting on the MS-COCO and Flickr30k datasets.
Our method shows promise for improvement using the cropping technique.
However, its performance still lags behind that of conventional methods.
Notably, our retrieval system successfully locates sentences containing exact matches to the query keyword based on a few words, as demonstrated in the keyword-based image retrieval evaluations.
Conversely, our system encounters difficulties in locating sentences that partially or ambiguously match the query caption sentence based on a combination of several words.
In the subsequent experiment, we provide a discussion of these results. 

\begin{table}[t!]
\begin{center}
\caption{Recall performance comparison between our retrieval method and the conventional retrieval methods in the caption-to-image retrieval setting using the MS-COCO and Flickr30k datasets.}
\scalebox{1.0}{
\begin{tabular}{p{8em}p{10em}p{10em}p{10em}|p{10em}p{10em}p{10em}}
\hline 
&\multicolumn{3}{c}{\textbf{MS-COCO}}&\multicolumn{3}{c}{\textbf{Flickr30k}}\\
&\multicolumn{3}{c}{5k validation set}&\multicolumn{3}{c}{1k test set}\\ \cline{2-7}
\multicolumn{1}{c}{\textbf{Method}}&\multicolumn{1}{c}{R@1}&\multicolumn{1}{c}{R@5}&\multicolumn{1}{c}{R@10}&\multicolumn{1}{c}{R@1}&\multicolumn{1}{c}{R@5}&\multicolumn{1}{c}{R@10}\\ \hline
\multicolumn{1}{c}{\textbf{ALIGN}~\cite{ALIGN}}&\multicolumn{1}{c}{40.2}&\multicolumn{1}{c}{64.5}&\multicolumn{1}{c}{74.7}&\multicolumn{1}{c}{81.4}&\multicolumn{1}{c}{96.7}&\multicolumn{1}{c}{98.7}\\
\multicolumn{1}{c}{\textbf{FLAVA}$^\dag$~\cite{FLAVA}}&\multicolumn{1}{c}{38.4}&\multicolumn{1}{c}{67.5}&\multicolumn{1}{c}{-}&\multicolumn{1}{c}{65.2}&\multicolumn{1}{c}{89.4}&\multicolumn{1}{c}{-}\\
\multicolumn{1}{c}{\textbf{CLIP}~\cite{CLIP}}&\multicolumn{1}{c}{33.9}&\multicolumn{1}{c}{58.5}&\multicolumn{1}{c}{69.2}&\multicolumn{1}{c}{73.6}&\multicolumn{1}{c}{93.4}&\multicolumn{1}{c}{97.3}\\
\multicolumn{1}{c}{\textbf{UNITER}$^\dag$~\cite{UNITER}}&\multicolumn{1}{c}{-}&\multicolumn{1}{c}{-}&\multicolumn{1}{c}{-}&\multicolumn{1}{c}{68.7}&\multicolumn{1}{c}{89.2}&\multicolumn{1}{c}{93.9}\\
\multicolumn{1}{c}{\textbf{ImageBERT}$^\dag$~\cite{ImageBERT}}&\multicolumn{1}{c}{32.3}&\multicolumn{1}{c}{59.0}&\multicolumn{1}{c}{70.2}&\multicolumn{1}{c}{54.3}&\multicolumn{1}{c}{79.6}&\multicolumn{1}{c}{87.5}\\ \hline
\multicolumn{1}{c}{\textbf{Ours}}&\multicolumn{1}{c}{22.1}&\multicolumn{1}{c}{42.5}&\multicolumn{1}{c}{53.2}&\multicolumn{1}{c}{42.8}&\multicolumn{1}{c}{67.4}&\multicolumn{1}{c}{75.3}\\
\multicolumn{1}{c}{\textbf{+crop}}&\multicolumn{1}{c}{27.3}&\multicolumn{1}{c}{49.5}&\multicolumn{1}{c}{59.9}&\multicolumn{1}{c}{57.3}&\multicolumn{1}{c}{81.8}&\multicolumn{1}{c}{88.0}\\ \hline
\multicolumn{7}{r}{$^\dag$ We refer to the values reported in the papers.}
\end{tabular}
}
\label{table:recall_comparison_caption}
\end{center}
\end{table}

\subsubsection{Text-to-image retrieval with user feedback.}

\begin{figure}[t!]
   \begin{center}
   \includegraphics[width=1.0\linewidth]{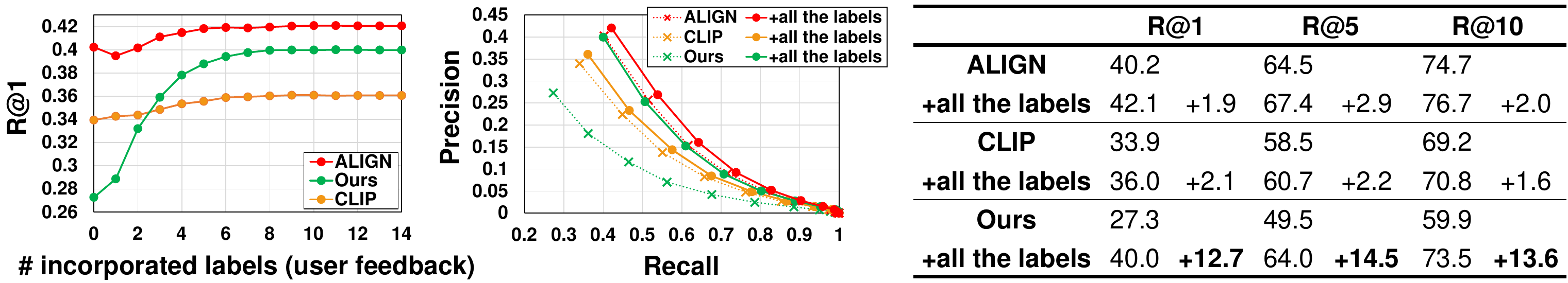} 
   \end{center}
   \caption{
   (Left) The variations of recall at 1 (R@1) for the text-to-image retrieval with user feedback setting on the MS-COCO dataset.
   The ground-truth labels are iteratively incorporated into the search query as the keyword-based user feedback after the initial retrieval based on the caption-to-image retrieval setting.
   (Center) The variations of precision and recall curves by incorporating all the ground-truth labels for each image into the search query.
   (Right) The summary of the improvement in recall performance with the keyword-based user feedback.}
   \label{fig:pr_curves_feedback_and_multi_turn}
\end{figure}

The left graph in Fig.~\ref{fig:pr_curves_feedback_and_multi_turn} demonstrates an improvement in recall (R@1) as the number of keyword-based user feedback increases,
by iteratively incorporating the ground-truth labels for each image in the MS-COCO dataset into the search query.
As shown in the center graph of Fig.~\ref{fig:pr_curves_feedback_and_multi_turn}, 
the precision at the similar recall is improved by incorporating all the ground-truth labels into the search query,
because images to be retrieved can be specified based on the additional keywords.
As an example, consider a caption sentence for an image file ``\texttt{val2017/000000003661.jpg}'', such as ``\texttt{A bunch of bananas sitting on top of a wooden table.}''
Our initial top-1 retrieval result for this query caption is an image file ``\texttt{val2017/000000571718.jpg.}''
This image actually depicts a bunch of bananas sitting on top of a wooden table, with a man standing nearby, which is relevant to the search query based on the caption.
In other words, the query caption is not sufficiently informative to precisely specify the desired image. 
If we incorporate the keywords ``\texttt{cup}'', ``\texttt{banana}'', and ``\texttt{keyboard}’’ into the search query based on the ground-truth labels of the image file ``\texttt{val2017/000000003661} \texttt{.jpg}'',
our system can successfully return the ground-truth image as the top-1 retrieved image.
In this case, the file ``\texttt{val2017/000000571718.jpg}'' is eliminated from the retrieved candidates since this image does not contain keyboards.

In another example, a caption sentence for an image file ``\texttt{val2017/00000000}\\ \texttt{2149.jpg}'' is ``\texttt{A large white bowl of many green apples.}''
Our initial top-1 retrieval result for this query caption is an image file ``\texttt{val2017/000000575970}\\ \texttt{.jpg}'',
which depicts a bowl of green apples on the dining table in the living room.
If we incorporate the keywords of ``\texttt{bowl}'' and ``\texttt{apple}'' into the search query based on the ground-truth labels of the image file ``\texttt{val2017/000000002149.jpg}'',
our system can successfully return the ground-truth image as the top-1 retrieved image by prioritizing the keywords in the query, thus specifying the focus points.

Finally, the right table in Fig.~\ref{fig:pr_curves_feedback_and_multi_turn} summarizes the quantitative improvement on recall performance.
When compared to the conventional methods, our retrieval method exhibits a significant improvement.

\section{Conclusions}

In this paper, we introduced an image retrieval system that utilizes an M-LLM to extract image features into textual data and that employs an efficient sparse retrieval algorithm commonly used in NLP tasks.
We considered the keyword-based image retrieval scenarios as text-to-image retrieval tasks, where keywords are utilized for search queries and refining the search criteria.
In the keyword-based image retrieval scenarios, we demonstrated that our approach outperforms the conventional vision-language model-based methods in terms of precision and recall on the benchmark datasets.
In particular, we introduced a cropping technique that assists the M-LLM in effectively extracting image features.
We analyzed the impact of the cropping technique by using CLIPScore, and empirically showed the effectiveness based on the improvement of the retrieval performance.
Finally, we demonstrated that the iterative incorporation of keywords into search queries like user feedback significantly improves our retrieval performance.

%
%
\clearpage

\bibliographystyle{splncs04}
\bibliography{main}

\newpage
\appendix

\section*{Appendix}

\section{Informativeness of Queries}

When a user searches for an image, the query may not always adequately represent or reflect the user's desires or intentions. It could be incomplete or lacking in the necessary information to specify them.
For example, consider a situation where a user is searching for an image that depicts a Labrador Retriever lying on a grass field.
If the search query is simply ``\texttt{Labrador Retriever},'' it lacks the necessary information to specify the desired images and a large number of Labrador Retriever images could be potential retrieval candidates.
In contrast, by combining keywords like ``\texttt{Labrador Retriever},'' ``\texttt{lying},'' and ``\texttt{grass field}'' in the search query, it becomes more informative and helps to specify and retrieve the desired image.
Experimental examples are presented in the experiments of text-to-image retrieval with user feedback in Sec.4.2 of the paper.

\section{Analysis with CLIPScore}

In Sec. 3.1 of the paper, we evaluate the effectiveness of cropping images using CLIPScore.
To calculate CLIPScore, we utilize images from the MS-COCO dataset and adopt the list of 80 categories of the MS-COCO dataset as diverse textual descriptions
(e.g., ``\texttt{bicycle}'' and ``\texttt{refrigerator}'').

\begin{figure}[t!]
   \begin{center}
   \includegraphics[width=1.0\linewidth]{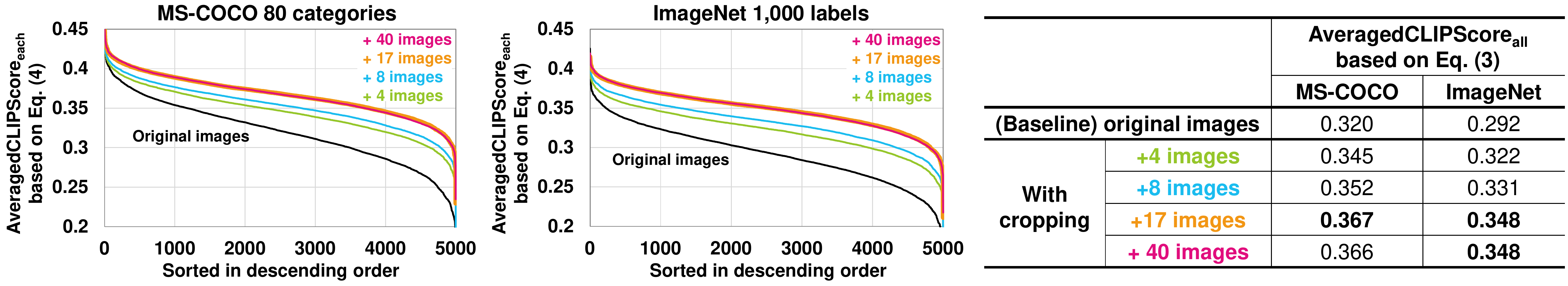}
   \end{center}
   \caption{The variations in averaged CLIPScore based on Eq. (4) of the paper for each of the 5,000 validation images from the MS-COCO dataset.
   We adopt the list of 80 categories of the MS-COCO dataset (left) and the list of 1,000 labels of the ImageNet dataset (center) as diverse textual descriptions.
   The left graph is same as in Fig. 3 of the paper.
   The right table summarizes averaged CLIPScore based on Eq. (3) of the paper for all the images in the MS-COCO dataset.
    }
   \label{fig:clipscore_imagenet}
\end{figure}

We can also adopt the other textual descriptions, such as a list of 1,000 labels in the ImageNet dataset~\cite{ImageNet},
which are not directly related to the images in the MS-COCO dataset.
We calculate CLIPScore between each image in the MS-COCO dataset and each text label of the ImageNet dataset.
We average the values based on Eqs. (3) and (4) of the paper,
and summarize the variations of the averaged CLIPScore for each image in Fig.~\ref{fig:clipscore_imagenet}.
The left and center graphs in Fig.~\ref{fig:clipscore_imagenet} exhibit similar trends and characteristics.
For example, as the number of cropped images increases to 17, the averaged CLIPScore based on Eq. (3) for all the images in the dataset also increases.
Additionally, when the number of cropped images increases from 17 to 40 by cropping with overlaps, the averaged CLIPScore based on Eq. (3) does not increase.
This indicates that the impact of cropping has reached a saturation point.
These findings are consistent with those reported in the paper, which supports the validity of the CLIPScore analysis using the list of 80 categories from the MS-COCO dataset as diverse textual descriptions in Sec. 3.1 of the paper.

\begin{figure}[b!]
   \begin{center}
   \includegraphics[width=1.0\linewidth]{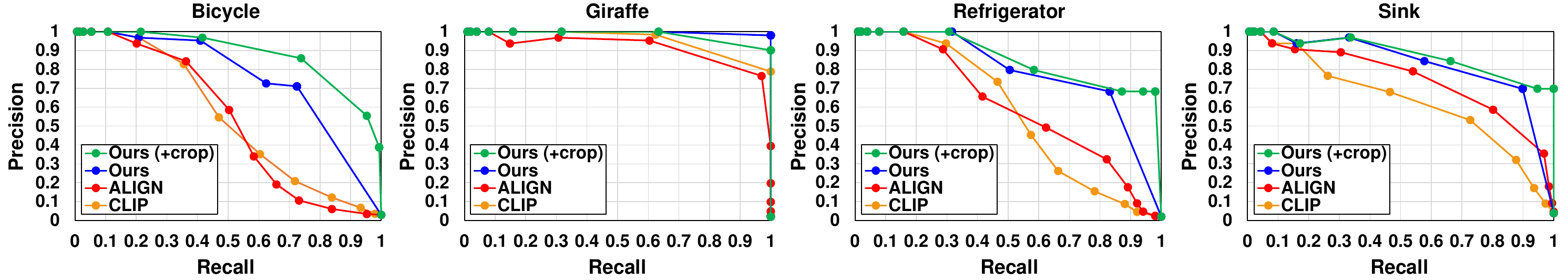}
   \end{center}
   \caption{Comparison of precision and recall curves for each category between our retrieval method and the conventional retrieval methods in the keyword-based image retrieval setting: (from left to right) bicycle, giraffe, refrigerator, sink.}
   \label{fig:pr_curves_each_category}
\end{figure}

\section{Precision and Recall Curves for Each Category}
\label{sec:precision_recall_curves_each_category}

We present the precision and recall curves for each category in the keyword-based image retrieval setting on the MS-COCO dataset as shown in Fig.~\ref{fig:pr_curves_each_category}.
As examples, we exhibit the curves for the categories of bicycle, giraffe, refrigerator, and sink. 
The characteristics vary depending on the categories, indicating that the ease of retrieving images differs across categories.
Overall, our method achieves higher precision and recall compared to the conventional methods,
and the performance is improved by using the cropping technique.
However, if the precision and recall are sufficiently high like the category of giraffe,
the cropping technique may not improve the performance and could potentially degrade the precision by retrieving irrelevant cropped images.

\section{Examples of Captions Generated by M-LLM}
\label{sec:example_generated_caption}

\begin{figure}[t!]
   \begin{center}
   \includegraphics[width=1.0\linewidth]{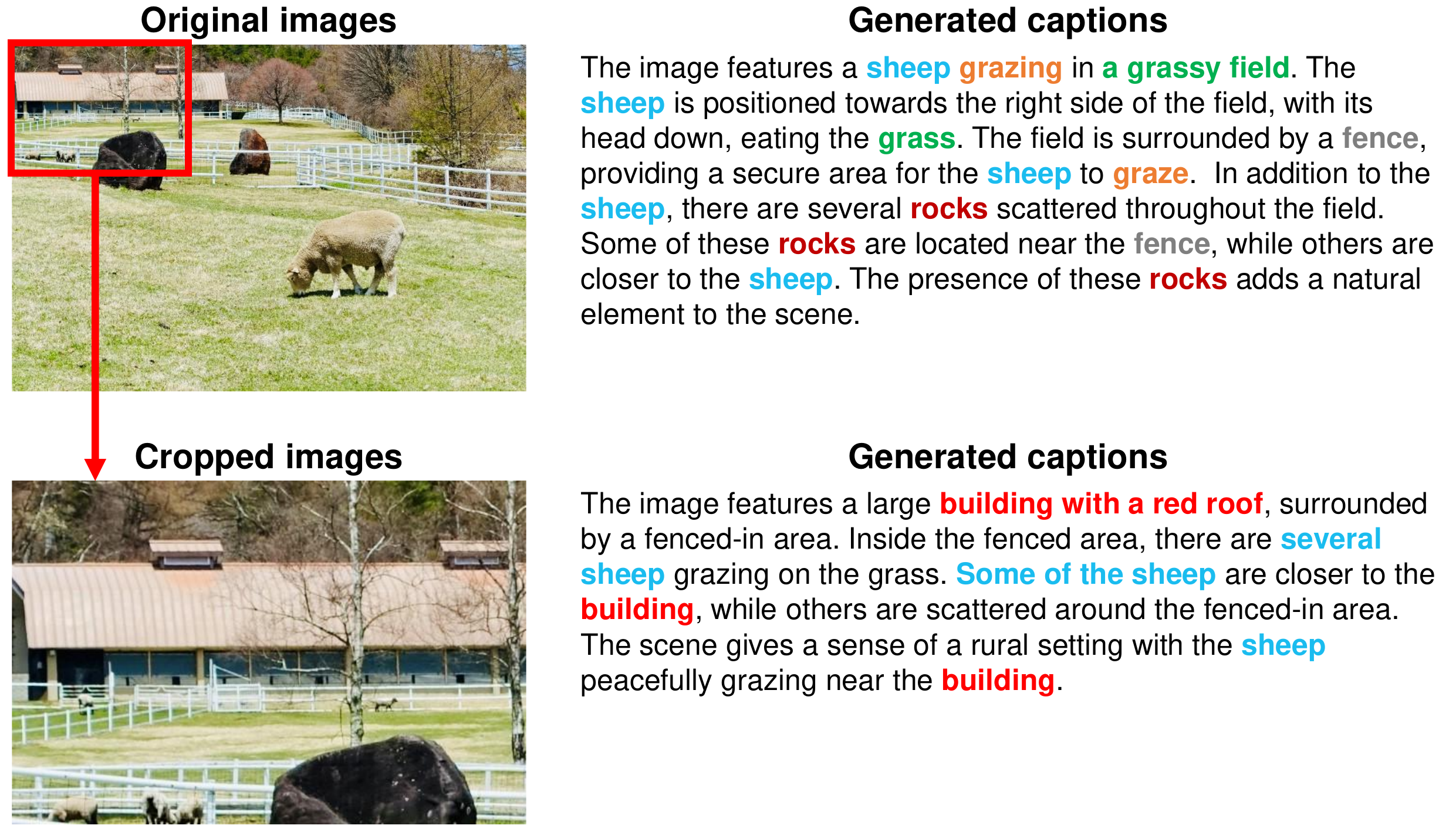}
   \end{center}
   \caption{Examples of captions generated by an M-LLM for an original image (top) and one of the cropped images (bottom). The repeated representations in the generated captions are intentionally highlighted in color.}
   \label{fig:example_generated_caption}
\end{figure}

Our system utilizes an M-LLM to generate textual data, such as tags and captions, that capture the semantic content of images.
As an example, captions generated by an M-LLM and corresponding images are shown in Fig.~\ref{fig:example_generated_caption}.
To generate the captions,
we employ the LLaVA's pre-trained model (\texttt{llava-1.5-13b-hf}) and provide the model with an image and a prompt ``\texttt{Please generate multiple captions to describe the features of this image.}'',
as described in Sec. 4.1 of the paper.
In this case, the words such as ``\texttt{sheep}'', ``\texttt{fence}'', and ``\texttt{rocks}'' frequently appeared in the generated captions.
Moreover, Fig.~\ref{fig:example_generated_caption} displays the generated captions for one of the cropped images.
By using the cropping technique, the M-LLM effectively extracts features from the image and reflects them in textual data.
For example, the phrase like ``\texttt{building with a red roof}'' appeared in the generated captions for the cropped image but not in those for the original image.
Furthermore, the M-LLM focuses on several sheep other than the one present in the center of the original image and incorporates this information into the generated captions. 

\section{Semantic Interpretability and Analyzability}

\begin{figure}[t!]
   \begin{center}
   \includegraphics[width=1.0\linewidth]{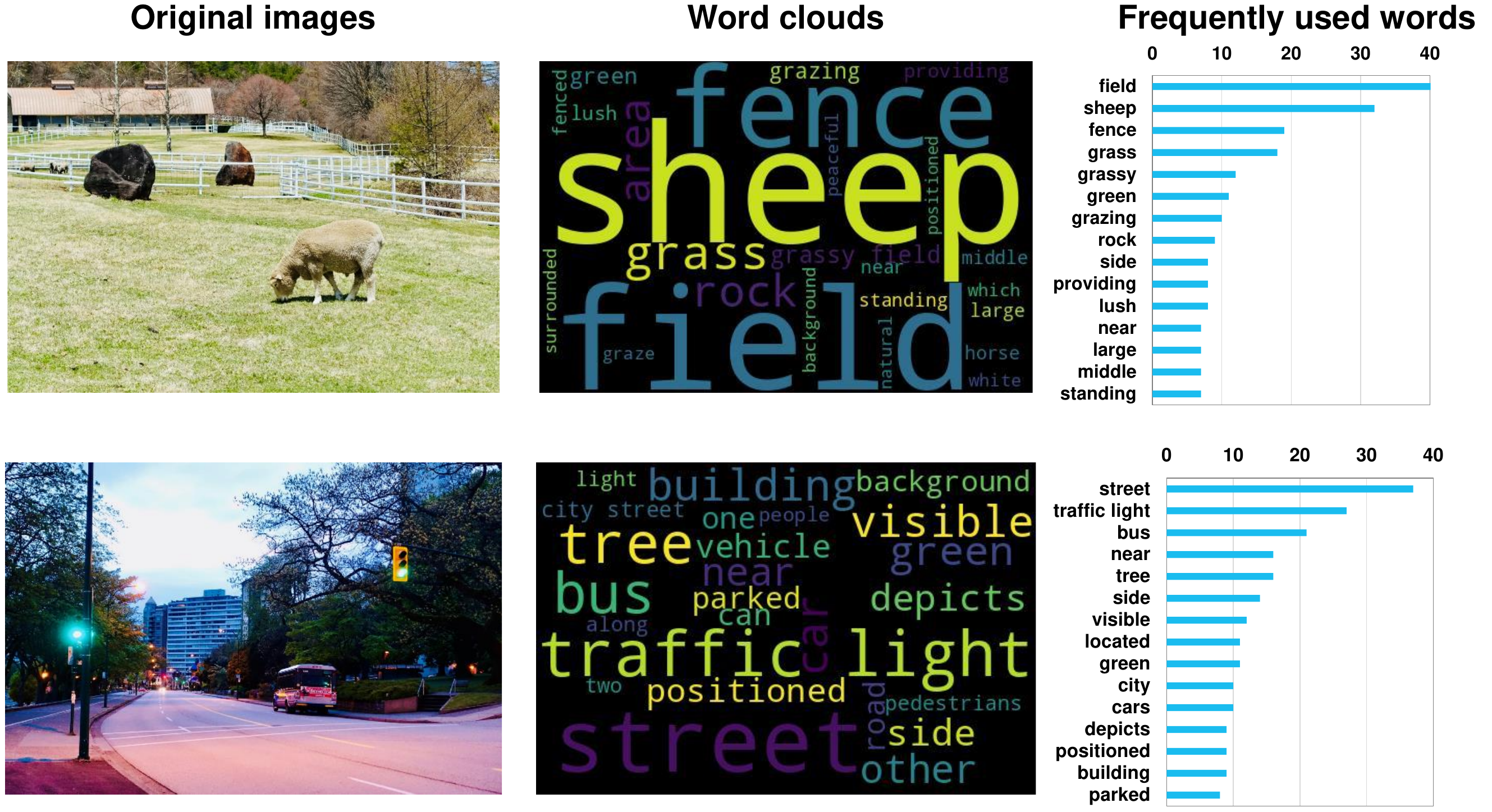}
   \end{center}
   \caption{Word cloud examples: the original images (left), word clouds based on the captions generated by M-LLMs (center), and histograms of the top 15 frequently used words in each image.}
   \label{fig:word_cloud}
\end{figure}

By viewing textual data generated by M-LLMs rather than encoded vector values, the semantic interpretability of images for humans can be enhanced.
Additionally, analyzing the textual data enables us to perform statistical analysis on image features.
For example, as shown in Fig.~\ref{fig:word_cloud},
we can create word clouds and histograms of the top 15 frequently used words based on the captions generated by the LLaVA's pre-trained model as described in Sec~\ref{sec:example_generated_caption}.
We use the cropping technique and concatenate all the generated captions for the original image and cropped images.
As shown in Fig.~\ref{fig:word_cloud}, the constituent elements and information present in images can be visualized and statistically analyzed.

Semantic segmentation techniques produce a pixel-wise segmentation map of an image, where each pixel is assigned to a specific class or object~\cite{SegNet,FCN,MaskRCNN}.
By counting pixels within segmented areas for each class or object, it also becomes possible to statistically analyze the constituent elements and information present in images.
In contrast, our approach leverages the representation capabilities of M-LLMs to extract diverse information from images, 
which provides a more comprehensive understanding of the image content compared to the traditional semantic segmentation methods that typically focus solely on predefined classes or object categories.

\section{Discussion for Limitations and Potential Negative Impact}

Sec. 3.1 of the paper describes one limitation of our approach,
which is its applicability mainly to images that can be described in language, such as scenic views containing objects.
Our approach utilizes M-LLMs that support visual prompting to extract features from images and represents them using textual descriptions.
Consequently, our approach might not be well-suited for images that are difficult to describe in language, such as medical images (e.g., chest x-rays) or defects in anomaly detection tasks.
In such cases, DNNs would need to be trained directly on the raw pixel values of the images and extract the image features in a latent space without relying on linguistic descriptions.
However, a text-to-image generation technique, called textual inversion, have been proposed to capture visual concepts in given images while keeping text-to-image models frozen~\cite{textual_inversion}.
This technique enables us to obtain vectors representing specific visual concepts in the latent space of the frozen text-to-image model.
These vectors can then serve as queries or keys in image retrieval tasks, even though the image features themselves cannot be described in language.
As part of our future work, we plan to further explore the combination of image features that can and cannot be described using language for image retrieval by leveraging such techniques.

One potential negative impact of our work is that if our approach is processed on edge devices,
it could significantly affect the limited battery life of these devices due to the substantial computational resources and energy consumption required by M-LLMs to extract image features.
If images stored locally on the edge devices can be synchronized with cloud backups,
the computationally intensive processes can be conducted in the cloud during periods of low usage.
For instance, if a user takes photographs during the daytime and approves them to be stored in the cloud for backups, the feature extraction process can be conducted in the cloud while the user is sleeping.
Once the feature extraction processes are complete, the resulting textual data can be downloaded to the user's device at a lower energy cost compared to using M-LLMs directly on the edge device, thereby saving the battery life.

\end{document}